\documentclass[10pt,twocolumn,letterpaper]{article}

\usepackage{cvpr}              % To produce the CAMERA-READY version
\definecolor{cvprblue}{rgb}{0.21,0.49,0.74}
\usepackage[pagebackref,breaklinks,colorlinks,allcolors=cvprblue]{hyperref}
\definecolor{tabhighlight}{RGB}{240,248,255}
\definecolor{top_panel}{HTML}{F4F9F9}
\definecolor{bottom_panel}{HTML}{FCF9F7}
\definecolor{cues}{HTML}{70AD47}
\definecolor{RL}{HTML}{579EBB}
\definecolor{tsne_blue}{HTML}{A6DADC}
\definecolor{tsne_green}{HTML}{A9D18E}
\usepackage{booktabs}
\usepackage{multirow}
\usepackage{rotating}
\usepackage[table]{xcolor}
\usepackage{graphicx}
\usepackage[normalem]{ulem}
\usepackage{arydshln}   % 为了支持 \cdashline

\usepackage{algorithm}
\usepackage{algpseudocode}
\usepackage{amsmath}
\usepackage{amssymb}
\def\paperID{3882} % *** Enter the Paper ID here
\def\confName{CVPR}
\def\confYear{2026}

\title{Incentivizing Generative Zero-Shot Learning via Outcome-Reward Reinforcement Learning with Visual Cues}

\author{Wenjin Hou$^{1}$\qquad Xiaoxiao Sun$^{2}$\qquad Hehe Fan$^{1,\, 3}$\thanks{Corresponding author.}\\
\textsuperscript{1}CCAI, Zhejiang University \qquad \textsuperscript{2}Stanford University \\
\textsuperscript{3}State Key Laboratory of CAD\&CG, Zhejiang University \\
{\tt\small houwj17@gmail.com $\quad$ \tt\small xxsun@stanford.edu $\quad$ hehefan@zju.edu.cn}
}

\begin{document}
\maketitle
\begin{abstract}
Recent advances in zero-shot learning (ZSL) have demonstrated the potential of generative models. Typically, generative ZSL synthesizes visual features conditioned on semantic prototypes to model the data distribution of unseen classes, followed by training a classifier on the synthesized data.
However, the synthesized features often remain task-agnostic, leading to degraded performance.
Moreover, inferring a faithful distribution from semantic prototypes alone is insufficient for classes that are semantically similar but visually distinct.
To address these and advance ZSL, we propose RLVC, an outcome-reward reinforcement learning RL framework with visual cues for generative ZSL.
At its core, RL empowers the generative model to self-evolve, implicitly enhancing its generation capability.
In particular, RLVC updates the generative model using an outcome-based reward, encouraging the synthesis of task-relevant features.
Furthermore, we introduce class-wise visual cues that (i) align synthesized features with visual prototypes and (ii) stabilize the RL training updates. 
For the training process, we present a novel cold-start strategy.
Comprehensive experiments and analyses on three prevalent ZSL benchmarks demonstrate that RLVC achieves state-of-the-art results with a 4.7\% gain.
%All code, models, and complete features will be open-sourced upon publication to accelerate future research.

\end{abstract}

\section{Introduction}
\label{sec:intro}

\begin{figure}
    \centering
    \includegraphics[width=1.0\linewidth]{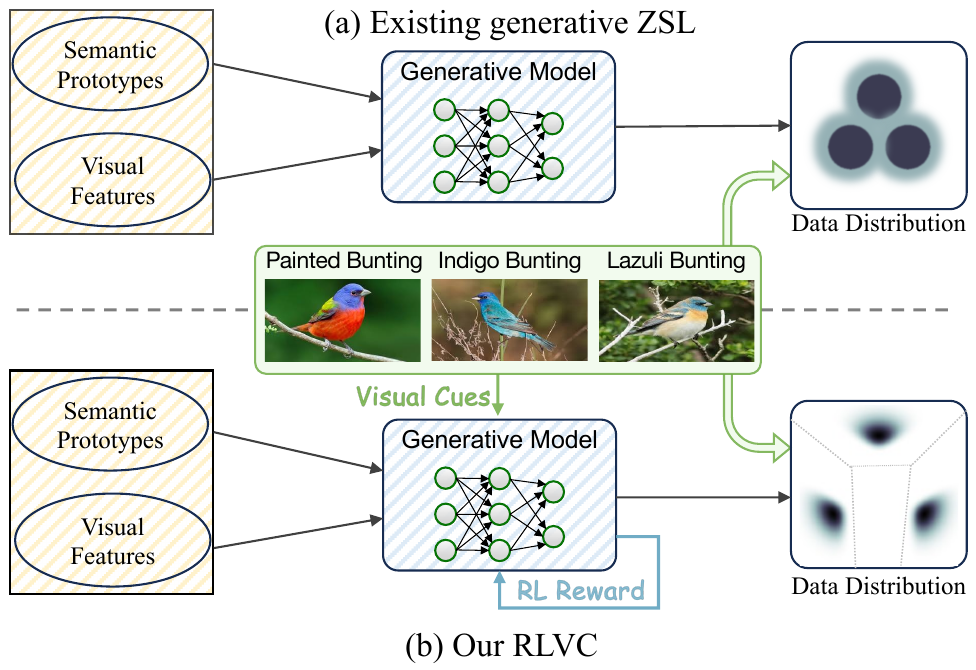}
    \vspace{-7mm}
    % \caption{Motivating illustration. (a) Existing generative ZSL methods optimize the model via adversarial losses
    % conditioned solely on semantic prototypes, which is task-agnostic and overlooks the differences between classes that are semantically similar but visually distinct. (b) Our RLVC incentivizes the generative model updating through an outcome-based reward and visual cues (serving as visual prototypes), enabling synthesized features that remain task-relevant and faithfully represent the data distribution.}
    \caption{Motivating illustration. (a) Existing generative ZSL methods train with adversarial losses conditioned only on semantic prototypes. This often leads to task-agnostic synthesized features and inter-class overlap. (b) Our RLVC incentivizes the generative model updating via RL reward and visual cues, enabling synthesized features that remain task-relevant and faithfully represent the data distribution.}
    \label{fig:motivation}
    \vspace{-4mm}
\end{figure}

Generative models (\eg, variational autoencoders (VAEs)) \cite{kingma2013auto}, generative adversarial networks (GANs) \cite{goodfellow2014generative}, and diffusion models (DMs) \cite{ho2020denoising}) have emerged as practical solutions for zero-shot learning (ZSL) \cite{xian2018feature, cetin2022closed, Hou_2024_CVPR}. 
Using predefined semantic prototypes (\eg, expert-annotated attribute vectors \cite{yan2021zeronas, chen2023evolving, hou2024zeromamba, ye2025zerodiff} or word embeddings of class names \cite{naeem2022i2dformer, wang2023improving, naeem2024i2dformer+, chen2025genzsl}) as conditions, these models synthesize high-quality visual features \cite{kong2022compactness, Hou_2024_CVPR, chen2025semantics} or images \cite{clark2023text, fu2024discriminative, kwon2024zero, hong2024galot} for unseen classes. 
As a result, they offer an unbounded representation space for modeling data distributions in generative ZSL, effectively alleviating the lack of unseen classes.

To improve synthesized samples and advance generative ZSL, existing ZSL solutions mainly pursue four directions: 
{\bf i}) fine-tuning the visual backbone on the seen classes of the ZSL dataset \cite{xian2019f,narayan2020latent,ye2025zerodiff}. 
{\bf ii}) enforcing visual–semantic consistency via bidirectional mapping \cite{hong2022semantic, li2023vs}. 
{\bf iii}) learning enhanced visual features for classifier training \cite{chen2021free, kong2022compactness, chen2022msdn, chen2025semantics}. 
{\bf iv}) evolving semantic prototypes with visual features for better alignment \cite{chen2023evolving, Hou_2024_CVPR, ye2025zerodiff,li2023dr}. 
Rather than aligning visual and semantic features from scratch, recent approaches, such as CLIP \cite{radford2021learning} and SHIP \cite{wang2023improving}, which utilize large-scale vision-language pre-training, also provide suitable class prototypes.  
Although impressive, the synthesized features obtained via these methods exhibit two limitations: {\bf Firstly}, the generative model is typically optimized independently of the downstream classifier, which restricts its ability to model task-relevant data distributions.
{\bf Secondly}, some methods rely exclusively on the semantic conditions, leading to overlapping inter-class feature distributions and misclassifications.
For example, the classes ``Indigo Bunting'', ``Lazuli Bunting'' and ``Painted Bunting'' are semantically similar but visually distinct (see Fig. \ref{fig:motivation}). Hence, we posit that richer supervision is necessary to bolster tolerance to inter-class variance.

In light of the above, we present \textbf{RLVC}, an outcome-reward reinforcement learning (RL) framework with class-wise visual cues for generative ZSL.
At its core, RL mimics human trial-and-error learning to achieve goals through self-evolution (\ie, learning how to take actions to maximize reward) \cite{shakya2023reinforcement, liu2025visual}.
Based on these advantages, we consider RL's outcome-based optimization well suited for improving the capability of generative models (the {\em ``Why''}).
Moreover, RL can align the generation process more directly with the downstream classification objective, rather than relying on adversarial losses tied only to semantic prototypes, thereby balancing inter-class separation and task relevance.
In addition, class-wise visual cues act as reliable supervision during the training process.

Specifically, we treat the generator (denoted as $G_\theta$) as a policy model from the view of RL \textcolor{red}{\footnotemark[1]}\footnotetext[1]{Throughout the paper, ``generative model”, ``generator”, and ``policy model” are interchangeable terms in our method.}.
For task-relevant generation, we design a classifier to serve as the reward model. 
We pre-train this reward model to produce an outcome-based score (\ie, the predicted probability of the given class) as the reward signal. This reward drives updates of $G_\theta$, explicitly aligning synthesized features with the downstream classification task.
Meanwhile, we mine class-wise visual cues from fine-tuned visual features of seen classes and take them as visual prototypes.
Then, we impose a prototype-distillation loss to directly align synthesized features with these visual prototypes.
As an additional benefit, visual cues also stabilize RL optimization. 

In terms of the training paradigm, we adopt a novel cold-start strategy. %(the {\em ``When''}).
We first perform some iterations using the generative loss, then activate RL training once a preset threshold is reached. 
To alleviate gradient conflict and improve optimization stability, we employ an alternating mechanism at each iteration (\ie, we update $G_\theta$ separately with the generative adversarial loss and the RL loss).
In addition, we fine-tune the visual features to mitigate cross-dataset bias and enhance generalization.
Consequently, RLVC synthesizes features that faithfully represent the data distribution and remain task-relevant (Fig. \ref{fig:motivation}(b)), effectively alleviating the above issues.

To sum up, our core contributions can be concluded as:
\begin{itemize}
    \item {\bf Novel perspective}. We present RLVC from an RL viewpoint. 
    To our knowledge, this is the first attempt to analyze and apply RL to generative ZSL.
    \item {\bf Controlled framework}. We introduce {\em an outcome-reward reinforcement-learning framework with visual cues} to incentivize the generative model. We further present a novel cold-start training recipe to stabilize the optimization process.
    \item {\bf Empirical validation}. We conduct systematic empirical validation across three critical ZSL benchmark datasets. Our experiments reveal that RLVC significantly {\em outperforms} current state-of-the-art (SOTA) ZSL methods.

\end{itemize}

    % \item {\bf Novel perspective}. We propose RLVC from an RL perspective. This work is the first to provide a principle for understanding {\em why and when} RL succeeds in generative ZSL. We hope that RLVC will serve as a foundation for future research in the ZSL realm.
\section{Related Works}
\label{sec:formatting}
\begin{figure*}[!t]
    \centering
    \includegraphics[width=1.0\linewidth]{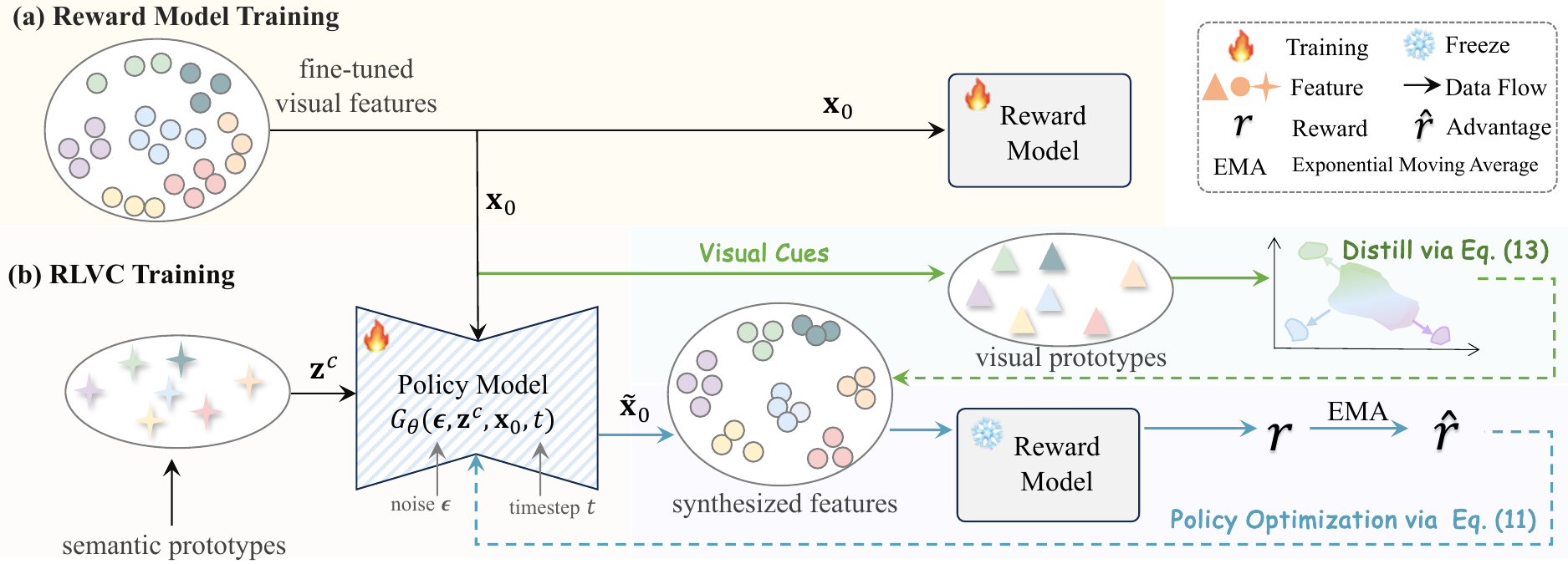}
    \vspace{-5mm}
     \caption{Model architecture and training of RLVC. The top panel shows how we train the reward model with a visual encoder to produce fine-tuned visual features and reward signals. The bottom panel depicts how we update the policy model $G_\theta$ (\ie, generator) via outcome-reward reinforcement learning ({\color{RL}blue arrows}) and visual cues ({\color{cues}green arrows}), enabling synthesized features that remain task-relevant and faithfully represent the data distribution. $\mathbf{x}_{0}$ and $\tilde{\mathbf{x}}_{0}$ denote the real and synthesized features of seen classes, respectively.} %Other symbols are defined in \S\ref{Methodology}.}
    \label{fig:framework}
    \vspace{-2mm}
\end{figure*}

\subsection{Zero-Shot Learning}
The goal of ZSL is to recognize unseen classes by knowledge learned from seen ones \cite{larochelle2008zero,palatucci2009zero}. Classical solutions are typically categorized into two paradigms according to the ultimate classification space.
The earlier one is projection methods (\ie, embedding methods), which directly map visual features to the semantic space with a transformation function supervised by semantic prototypes \cite{xie2019attentive,huynh2020fine, wang2021dual, xu2022attribute, chen2023duet,chen2022zero, naeem2023i2mvformer, chen2023zero, naeem2024i2dformer+, duan2024visual,chen2024causal,hou2024zeromamba, liu2025attend, chen2025dynamic, atigh2025simzsl}. 
The second paradigm can be viewed as a data augmentation strategy (\ie, generative methods). 
It synthesizes visual features to represent the data distribution of unseen classes. 
Then, training a classifier in the visual space to perform ZSL \cite{xian2018feature, han2021contrastive, chen2021free,chen2021hsva,yan2021zeronas,cetin2022closed,chen2023evolving,chen2023egans,su2023dual,10158446,Hou_2024_CVPR,fu2024discriminative,ye2025zerodiff,chen2025semantics}.
For embedding methods, recent advances mainly learn locally aligned visual--semantic correspondences via attention mechanisms.
For instance, TransZero++ \cite{chen2022transzero,chen2022transzero++} and ZSLViT \cite{chen2024progressive} employ transformer-based cross-modal attention to align visual patches with attributes.
PSVMA+ \cite{liu2023progressive,liu2024psvma+} and VSPCN \cite{jiang2025visual} enforce progressive visual--semantic mutual alignment.
Despite these advancements, the learned embedding space of this line of work often exhibits bias toward seen classes \cite{ye2023rebalanced}. 
Besides, under weak semantic conditions (\eg, class names), achieving fine-grained local alignment remains challenging \cite{chen2025genzsl,xiang2026when}. 
Therefore, generative methods have received increasing attention recently.

Within the generative paradigm, early methods focus on bidirectional semantic$\leftrightarrow$visual consistency with simple decoders and $\ell_1/\ell_2$ losses, such as cycle-CLSWGAN \cite{felix2018multi}, FREE \cite{chen2021free}, and TFVAEGAN \cite{narayan2020latent}. 
LisGAN \cite{li2019leveraging} and  LsrGAN \cite{vyas2020leveraging} take a classification loss as a part of the discriminator loss.
Subsequent work improves training strategies: CE-GZSL \cite{han2021contrastive} and ICCE \cite{kong2022compactness} learn contrastive embeddings; ZLAP \cite{chen2022zero} adjusts logits; ESZSL \cite{cetin2022closed} proposes sample probing, SC-EGG \cite{hong2022semantic} adopts an embedding-guided generator; and ZeroNAS \cite{yan2021zeronas} introduces NAS into ZSL. 
More recent advances, such as TDCSS \cite{feng2022non}, DSP \cite{chen2023evolving}, VADS \cite{Hou_2024_CVPR}, GenZSL \cite{chen2025genzsl}, ViFR \cite{chen2025semantics}, and ZeroDiff \cite{ye2025zerodiff}, inject stronger priors and model unseen class–conditional distributions more faithfully \cite{zheng2026vii}. 
Despite this progress, two structural issues persist: \textbf{i}) most approaches optimize the generator independently of the downstream classifier, yielding \emph{task-agnostic} features; and \textbf{ii}) semantic-only conditions often induce \emph{inter-class confusion}, especially for fine-grained categories that are semantically similar but visually distinct. 
Inspired by the success of RL in vision, we propose an RL framework to improve the generator's modeling capability.
Meanwhile, we mine class-wise visual cues to distill richer visual information for feature synthesis.

\subsection{Reinforcement Learning for Visual Tasks}
RL optimizes decision-making by interacting with an environment to maximize cumulative reward.
Its applications span gaming, embodied AI, finance, and puzzle \cite{ghasemi2024comprehensive, song2025visualpuzzles,pmlr-v267-tan25f,Shi_2025_ICCV}. 
More recently, RL has played an increasingly important role in post-training of large language models (LLMs) to enhance reasoning ability \cite{kumar2025llm}. 
Frontier models such as DeepSeek-R1 \cite{guo2025deepseek}, OpenAI o1 \cite{jaech2024openai}, and Qwen \cite{wang2024qwen2,yang2025qwen3} employ verifiable rewards to optimize task performance, encouraging models to align with human intentions \cite{wang2025internvl3,liu2025prorl}.

Motivated by RL's ability to self-evolve in the language domain and its strong generalization, the vision community has begun to explore effective RL frameworks. 
For example, Visual-RFT \cite{liu2025visual} and VPRL \cite{xu2025visual} introduce visual reinforcement fine-tuning (RFT) and extend RFT to visual tasks.
Broadly, these efforts either design task-grounded rewards or feedback (\eg, accuracy rewards, format rewards \cite{liu2025visual}) or develop more efficient optimization strategies (\eg, DPO \cite{xu2024dpo}, GRPO \cite{shao2024deepseekmath}).
Building on these insights, we make the first attempt to investigate whether RL can advance ZSL.
Rather than proposing complex policy optimization algorithms, we aim to provide a simple pipeline based on outcome-reward RL for faithful visual synthesis in ZSL.

\section{RLVC}
\label{Methodology}

 \noindent \textbf{Motivation and Overview.} 
 In this section, we present the core design of RLVC. 
 Our objective is to extend the generative model's modeling capacity to construct data representations well-suited for classification. 
 To this end,  we treat the generator as a policy and optimize it via RL. 
 A key component in RLVC is a reward model that delivers task-specific reward signals, enabling stronger model optimization. 
 Within a coherent framework, RLVC first trains the reward model together with the visual encoder. 
 During policy training, the fine-tuned visual features are used to mine visual cues, and the reward model produces an outcome reward (see Fig. \ref{fig:framework}).
 We detail these designs and the training procedure in $\S$\ref{rl}, $\S$\ref{vc} and $\S$\ref{training}.

\noindent \textbf{Problem Statement.} 
Let $\mathcal{Y}^{s}$ and $\mathcal{Y}^{u}$ denote the disjoint sets of seen and unseen classes, respectively, with $\mathcal{Y}^{s}\cap\mathcal{Y}^{u}=\emptyset$ and $C^{s}=|\mathcal{Y}^{s}|$, $C^{u}=|\mathcal{Y}^{u}|$. Training data are available only for seen classes: $\mathcal{D}^{tr}=\{(\mathbf{x}_{i}^{s},\,y_{i}^{s},\,\mathbf{z}^{\,y_{i}^{s}})\}_{i=1}^{N_{tr}}$, where $\mathbf{x}_{i}^{s}\in\mathcal{X}$ is a visual feature, $y_{i}^{s}\in\mathcal{Y}^{s}$ is its label, and $\mathbf{z}^{\,c}\in\mathcal{A}$ denotes the semantic prototype for class $c$ (\eg, an attribute vector or a text embedding). Semantic prototypes $\{\mathbf{z}^{\,c}\}_{c\in\mathcal{Y}^{s}\cup\mathcal{Y}^{u}}$ are assumed to be available for both seen and unseen classes. In the conventional ZSL (CZSL) setting, the test set contains only unseen classes $\mathcal{D}^{te,u}=\{(\mathbf{x}_{j}^{u},\,y_{j}^{u},\,\mathbf{z}^{\,y_{j}^{u}})\}_{j=1}^{N_{u}}$, and the goal is to learn a classifier $f_{\text{CZSL}}:\mathcal{X}\rightarrow\mathcal{Y}^{u}$. In the generalized ZSL (GZSL) setting, the test set includes both seen and unseen samples, $\mathcal{D}^{te}=\mathcal{D}^{te,s}\cup\mathcal{D}^{te,u}$ with $\mathcal{D}^{te,s}\subseteq \mathcal{X}\times\mathcal{Y}^{s}$ and $\mathcal{D}^{te,u}\subseteq \mathcal{X}\times\mathcal{Y}^{u}$, and the goal becomes $f_{\text{GZSL}}:\mathcal{X}\rightarrow\mathcal{Y}^{s}\cup\mathcal{Y}^{u}$. The objective in both settings is to leverage $\mathcal{D}^{tr}$ and the semantic prototypes $\{\mathbf{z}^{\,c}\}$ to minimize the expected classification error on the corresponding test domain.

\subsection{Diffusion-based Generative Framework}
Motivated by the strong generative capacity of diffusion models, we adopt a diffusion-based adversarial framework \cite{xiao2021tackling, ye2025zerodiff}. 
For simplicity, the framework comprises a generator \(G_{\theta}\) and two discriminators, \(D_{x_0}\) and \(D_{x_t}\).
Given a class prototype \(\mathbf{z}^{\,c}\), Gaussian noise \(\boldsymbol{\epsilon}\!\sim\!\mathcal{N}(0,\mathbf{I})\), and a diffusion state \(\mathbf{x}_{t}\) at timestep \(t\), the generator outputs a visual feature:
\begin{equation}
\tilde{\mathbf{x}}_{0}
\;=\;
G_{\theta}\!\big(\boldsymbol{\epsilon},\,\mathbf{z}^{\,c},\,\mathbf{x}_{t},\,t\big)
\,\in\,\mathbb{R}^{d}.
\label{eq:gen-output}
\end{equation}

Conditioned on $\mathbf{z}^{\,c}$, the two discriminators operate as follows.
$D_{x_0}$ distinguishes real clean features ($\mathbf{x}_{0}$) from synthesized ones ($\tilde{\mathbf{x}}_{0}$).
$D_{x_t}$ contrasts the real transition ($\mathbf{x}_{t}$, $\mathbf{x}_{t+1}$) with the synthesized transition ($\tilde{\mathbf{x}}_{t}$, $\mathbf{x}_{t+1}$), where $\tilde{\mathbf{x}}_{t}$ is obtained from $\tilde{\mathbf{x}}_{0}$ and $\mathbf{x}_{t+1}$ via a first-order posterior update.
The optimization objectives are as follows:
\begingroup
\small 
\begin{equation}
\mathcal{L}_{D_{x_0}} =
- \mathbb{E}\!\big[D_{x_0}(\mathbf{x}_{0},\mathbf{z}^{\,c})\big]
+ \mathbb{E}\!\big[D_{x_0}(\tilde{\mathbf{x}}_{0},\mathbf{z}^{\,c})\big]
+ \lambda_{\mathrm{gp}}\,\mathrm{GP}_{x_0},
\label{eq:disc-x0}
\end{equation}
\endgroup
\begin{equation}
\begin{split}
\mathcal{L}_{D_{x_t}} ={}&
-\mathbb{E}\!\big[D_{x_t}(\mathbf{x}_{t},\mathbf{x}_{t+1},\mathbf{z}^{\,c},t)\big] \\
&\quad + \mathbb{E}\!\big[D_{x_t}(\tilde{\mathbf{x}}_{t},\mathbf{x}_{t+1},\mathbf{z}^{\,c},t)\big]
+ \lambda_{\mathrm{gp}}\,\mathrm{GP}_{x_t},
\end{split}
\label{eq:disc-xt}
\end{equation}
\begin{equation}
\mathcal{L}_{D} = \mathcal{L}_{D_{x_0}} + \mathcal{L}_{D_{x_t}},
\label{eq:disc-total}
\end{equation}
\begin{equation}
\mathcal{L}_{G}^{\mathrm{adv}}
= -\mathbb{E}\!\big[D_{x_0}(\tilde{\mathbf{x}}_{0},\mathbf{z}^{\,c})\big]
  - \mathbb{E}\!\big[D_{x_t}(\tilde{\mathbf{x}}_{t},\mathbf{x}_{t+1},\mathbf{z}^{\,c},t)\big],
\label{eq:gen-adv}
\end{equation}
where \(\lambda_{\mathrm{gp}}\) is the gradient-penalty weight and $\mathrm{GP}$ denotes the gradient penalty. During training, we alternately update the discriminator objective in Eq.~\eqref{eq:disc-total} and the generator objective in Eq.~\eqref{eq:gen-adv}. %Architectural details are provided in the Appendix \textcolor{cvprblue}{A}. %\ref{app:diffusion-feature}.

\subsection{Outcome-Reward Reinforcement Learning}
\label{rl}
In generative ZSL, a core challenge is to align generative capacity with discriminative representations that downstream classifiers can reliably use. 
However, purely adversarial objectives often synthesize task-agnostic features. 
To bridge this gap, we introduce an \emph{outcome-reward reinforcement learning (RL)} to directly incentivize the generator toward task-relevant synthesis.
From the RL perspective, we optimize the policy via self-evolving updates to favor features that are more likely to be correctly classified.
Concretely, a frozen classifier $R$ serves as the reward model. $R$ is implemented as a linear layer, \ie, $R(\mathbf{x}) = W\mathbf{x} + \mathbf{b}$ with $W \in \mathbb{R}^{C \times d}$, mapping a $d$-dim visual feature to $C$ class logits.
Given a synthesized feature $\tilde{\mathbf{x}}_0$, $R$ outputs logits that are converted to probabilities via the $\rm softmax$ operation,
\begin{equation}
p(y \mid \tilde{\mathbf{x}}_0) \;=\; \mathrm{softmax} \!\big(R(\tilde{\mathbf{x}}_0)\big)_y,
\label{eq:softmax}
\end{equation}
and we calculate the log-probability of the ground-truth class as the outcome reward $r$:
\begin{equation}
r \;=\; \log p\!\left(y \mid \tilde{\mathbf{x}}_0\right).
\label{eq:reward}
\end{equation}

Intuitively, higher confidence of $R$ on the correct class for $\tilde{\mathbf{x}}_0$ yields a larger $r$, which in turn steers the generator’s updates accordingly.

Furthermore, to stabilize training and improve performance, we use an exponential moving average (EMA) baseline $b$ over mini-batch rewards to compute the advantage $\hat r_{i}$. This process is formally defined as:
\begin{equation}
b \;\leftarrow\; \alpha\, b \;+\; (1-\alpha)\,\frac{1}{B}\sum_{i=1}^{B} r_i,
\label{eq:ema}
\end{equation}
\begin{equation}
\hat r_i \;=\; r_i \;-\; b,
\label{eq:adv}
\end{equation}
where $B$ is the batch size and $\alpha\!\in[0,1)$ controls smoothing (in experiments, we set $\alpha~{=}~0.9$). 
Moreover, we employ a stop-gradient operator $\operatorname{sg}[\cdot]$ for $\hat r_i$, so that it is treated as a constant without gradients and define the final advantage as:
\begin{equation}
\widehat{A}_i=\operatorname{sg}[\hat r_i].
\label{eq:advantage}
\end{equation}

Finally, the RL objective $\mathcal{L}_{\mathrm{RL}}$ for the policy model is:
\begin{equation}
\mathcal{L}_{\rm RL}
\;=\;
-\,\frac{1}{B}\sum_{i=1}^{B}\,
\widehat{A}_i \;\log p\!\left(y_i \mid \tilde{\mathbf{x}}_{0,i}\right).
\label{eq:rl-loss}
\end{equation}

Since $\log p(y_i\mid\tilde{\mathbf{x}}_{0,i})$ is differentiable with respect to $\tilde{\mathbf{x}}_{0,i}$, gradients propagate through $R$ and Eq.~\eqref{eq:softmax} to $\tilde{\mathbf{x}}_{0,i}$ and then to $G_\theta$ (while keeping $R$’s parameters fixed).
This outcome-based reward drives the policy to enhance its generation capability with clear guidance, enabling more discriminative, task-relevant synthesis in the feature space.

\begin{figure*}[!t]
    \centering
    \includegraphics[width=1.0\linewidth]{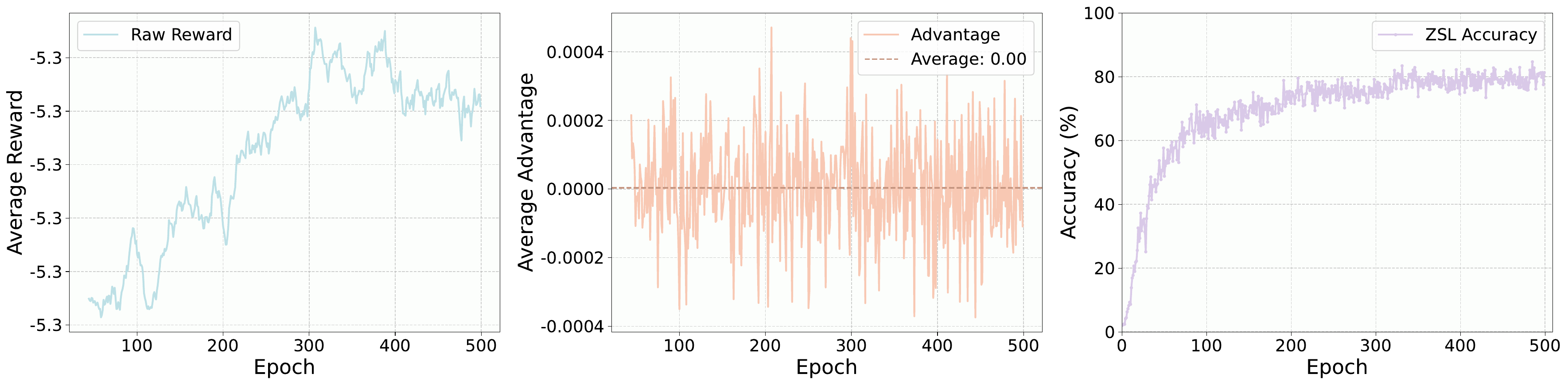}
    \vspace{-8mm}
    \caption{The training trends of our RLVC on CUB, including raw reward, EMA-adjusted advantage and ZSL accuracy.}
    \label{fig:training_trend}
    \vspace{-2mm}
\end{figure*}

\subsection{Visual Cues with Prototype-Distillation Loss}
\label{vc}
Although semantic prototypes provide strong conditions, they are inadequate to faithfully represent data distributions for classes that are semantically similar but visually distinct.
Additionally, it is common to include a Kullback–Leibler (KL) regularizer in the objective to constrain distributional shift and stabilize training in RL. 
Building on these insights, we introduce class-wise \emph{visual cues} as visual prototypes, following prototype learning frameworks \cite{qu2025learning}.
As illustrated in Fig. \ref{fig:framework}, we mine visual cues from the fine-tuned visual features.
Specifically, we gather all features belonging to each seen class and compute their mean. Formally, let $\mathcal{I}_c=\{\,i \mid y_i^s=c\,\}$ index the training samples of class $c\in\mathcal{Y}^s$, and let $\mathbf{x}_i^s\in\mathbb{R}^d$ denote the corresponding fine-tuned visual features. The visual prototype is:
\begin{equation}
\mathbf{v}^{\,c}
\;=\;
\frac{1}{|\mathcal{I}_c|}\sum_{i\in\mathcal{I}_c}\mathbf{x}_i^s \;\in\; \mathbb{R}^d.
\label{eq:proto}
\end{equation}

During policy optimization, we impose a prototype-distillation loss $\mathcal{L}_{\mathrm{PD}}$ that distills the synthesized features to align with the corresponding visual prototype:
\begin{equation}
\mathcal{L}_{\mathrm{PD}}=
\frac{1}{B}\sum_{i=1}^{B}\Bigg(1-
\frac{\tilde{\mathbf{x}}_{0,i}^{\top}\mathbf{v}^{\,c_i}}
{\|\tilde{\mathbf{x}}_{0,i}\|_2\,\|\mathbf{v}^{\,c_i}\|_2}\Bigg).
\label{eq:pd}
\end{equation}

This term pulls generated features closer to real data distribution and stabilizes the RL updates. 
We incorporate this loss into the generator update:
\begin{equation}
\mathcal{L}_{G}^{\text{total}}
\;=\;
\mathcal{L}_{G}^{\mathrm{adv}} \;+\; \lambda_{\mathrm{PD}}\,\mathcal{L}_{\mathrm{PD}},
\label{eq:G-total-pd}
\end{equation}
where $\lambda_{\mathrm{PD}}$ is a coefficient. We optimize $G_\theta$ by minimizing $\mathcal{L}_{G}^{\text{total}}$.
We also compare $\mathcal{L}_{\mathrm{PD}}$ with alternative losses (\eg, KL, $\ell_{\rm{1}}$) to validate the robustness of our design (Table \ref{table:ablation_loss}).

\subsection{Cold-Start Training Procedure}
\label{training}
To ensure effective optimization, RLVC adopts a novel cold-start schedule inspired by post-training frameworks for large language models (\ie, RL is disabled during the initial phase) \cite{guo2025deepseek}. 
Concretely, we first train for some epochs using the adversarial objective until the synthesized features exhibit basic class separability. Then, RL is activated to align generation with the downstream classification task further. 
To avoid gradient conflict after the cold-start threshold (\ie, $E_{\rm{RL}}$), we alternately update the policy model $G_\theta$ within each iteration via Eq.~\eqref{eq:G-total-pd} and Eq.~\eqref{eq:rl-loss}, rather than simply summing the two losses. 

We summarize the complete procedure in Algorithm~\ref{alg:rlvc} for clarity. We also report the training trends of the raw reward, the EMA-adjusted advantage, and ZSL accuracy in Fig. \ref{fig:training_trend}. 
Empirically, the reward increases and then stabilizes, the advantage exhibits only small fluctuations, and the ZSL accuracy gains steadily.
These trends consistently indicate that the model trains successfully and remains stable.

\begin{algorithm}[!t]
\caption{Training the Policy Model $G_{\theta}$}
\label{alg:rlvc}
\begin{algorithmic}[1]
\Require Minibatches $(\mathbf{x},y,\mathbf{z},\mathbf{v})$; weights $\lambda_{\mathrm{PD}}$; RL start epoch $E_{\mathrm{RL}}$; total epochs $E$; critic steps $K$.
\State Initialize $G_{\theta}$, $D_{x_0},D_{x_t}$; freeze reward model $R$.
\Procedure{Training}{$G_{\theta}$, $D_{x_0},D_{x_t}$}
\For{epoch $=1$ to $E$}
  \For{each minibatch}
    \For{$k=1$ to $K$} 
      \State Sample noise, timesteps $\tilde{\mathbf{x}}_{0}$ and $\tilde{\mathbf{x}}_{t}$.
      \State Update $D_{x_0},D_{x_t}$ via Eq.~\eqref{eq:disc-total}.
    \EndFor

    \State Sample noise, timesteps, $\tilde{\mathbf{x}}_{0}$ and $\tilde{\mathbf{x}}_{t}$.
    \State Compute $\mathcal{L}^{\mathrm{adv}}_{G}$ via Eq.~\eqref{eq:gen-adv}.
    \State Compute $\mathcal{L}_{\mathrm{PD}}$ via Eq. \eqref{eq:pd}.
    \State Update $G_{\theta}$ via Eq.\eqref{eq:G-total-pd}. \Comment{adversarial}

    \If{epoch $\ge E_{\mathrm{RL}}$} \Comment{cold-start threshold}
      \State Form advantage via Eqs.~\eqref{eq:softmax}--\eqref{eq:advantage}.
      \State Update $G_{\theta}$ via Eq.~\eqref{eq:rl-loss}. \Comment{RL update}
    \EndIf
  \EndFor
\EndFor
\State \Return optimized policy $G_{\theta}$
\EndProcedure
\end{algorithmic}
\end{algorithm}

\subsection{Inference for CZSL and GZSL}
At inference time, we freeze the generator $G_{\theta}$ and synthesize visual features for unseen classes via Eq.~\eqref{eq:gen-output}. No additional tricks are used. For CZSL, we train a standard softmax classifier solely on the synthesized unseen features. For GZSL, we train the classifier on the union of the finetuned seen features and the synthesized unseen features. Further details of the classifier follow prior work \cite{chen2023evolving, Hou_2024_CVPR}.

\section{Experiments}
In this section, we evaluate RLVC, aiming to answer the following questions:
\textit{(1) Can RLVC effectively and consistently improve ZSL accuracy on standard benchmarks (Tables \ref{tab:gzsl} and \ref{tab:semantic_prototype})?
(2) Do individual design choices of RLVC contribute to the observed accuracy gains (Tables \ref{table:ablation} and \ref{table:ablation_loss})?
(3) Does RLVC induce task-relevant feature distributions in the representation space (Fig. \ref{fig:tsne_cub})?
(4) How do hyperparameters impact performance (Fig. \ref{fig:hyper_cub})?}
%Before that, we first present the experimental setup.

\begin{table*}[!t]
  \caption{
  Compared our RLVC with the SOTA methods in CZSL and GZSL settings on CUB, SUN and AWA2 benchmarks. The symbol  ``$\star$" indicates the semantic prototypes from the class name.
  The symbol ``–" denotes that no results are provided in the original papers.
  The \textbf{bold} and \underline{underlined} markings indicate the best and second-best results, respectively. 
  }
  \label{tab:gzsl}
  \centering
  \resizebox{1.0\textwidth}{!}{
  \begin{tabular}{l@{\hspace{1.2mm}}rr|c|cccc|cccc|cccc}
    \toprule
    \multicolumn{3}{r}{\multirow{2}[2]{*}{\textbf{Method}}} & 
    \multicolumn{1}{c}{\multirow{2}[2]{*}{\textbf{Backbone}}} & 
    \multicolumn{4}{c|}{CUB} & 
    \multicolumn{4}{c|}{SUN} & 
    \multicolumn{4}{c}{AWA2} \\
    \cmidrule(rl){5-8} \cmidrule(rl){9-12} \cmidrule(rl){13-16}
    \multicolumn{3}{c|}{} & & 
    \textbf{Acc} & \textbf{U} & \textbf{S} & \textbf{H} & 
    \textbf{Acc} & \textbf{U} & \textbf{S} & \textbf{H} & 
    \textbf{Acc} & \textbf{U} & \textbf{S} & \textbf{H} \\
    \midrule

    \multirow{12}[2]{*}{\rotatebox{90}{Embedding}}
      & \multicolumn{2}{r|}{CLIP $^\star$ \textcolor{gray}{[ICML'21]} \cite{radford2021learning}} & ViT 
      & -- & 55.2  & 54.8 & 55.0  
      & -- & --  & --  & --  
      & -- & --  & --  & --  \\
      
      % & \multicolumn{2}{r|}{CoOP \textcolor{gray}{[IJCV'22]} \cite{zhou2022learning}} & ViT 
      % & -- & 49.2  & 63.8 & 55.6  
      % & -- & --  & --  & --  
      % & -- & 72.7  & \underline{95.3}  & 82.5  \\

      & \multicolumn{2}{r|}{TransZero++ \textcolor{gray}{[TPAMI'22]} \cite{chen2022transzero++}} & ResNet 
      & 78.3 & 67.5  & 73.6  & 70.4  
      & 67.6 & 48.6  & 37.8  & 42.5  
      & 72.6 & 64.6  & 82.7  & 72.5 \\

      & \multicolumn{2}{r|}{DUET \textcolor{gray}{[AAAI'23]} \cite{chen2023duet}} & ViT 
      & 72.3 & 62.9 & 72.8 & 67.5  
      & 64.4 & 45.7 & 45.8 & 45.8  
      & 69.9 & 63.7 & 84.7  & 72.7  \\

      & \multicolumn{2}{r|}{ICIS \textcolor{gray}{[ICCV'23]} \cite{christensen2023image}} & ResNet 
      & 60.6 & 45.8 & 73.7  & 56.5  
      & 51.8 & 45.2  & 25.6  & 32.7  
      & 64.6 & 35.6  & \underline{93.3}  & 51.6  \\

      & \multicolumn{2}{r|}{HAS \textcolor{gray}{[ACM MM'23]}  \cite{chen2023zero}} &ResNet 
      & 76.5& 69.6& 74.1& 71.8
      & 63.2& 42.8& 38.9& 40.8
      & 71.4& 63.1& 87.3& 73.3\\

      % & \multicolumn{2}{r|}{I2MVFormer \textcolor{gray}{[CVPR'23]} \cite{naeem2023i2mvformer}} & text document 
      % & -- & 42.5 & 59.9  & 49.7  
      % & -- & --  & --  & --  
      % & -- & 75.7  & 79.6  & 77.6  \\

      & \multicolumn{2}{r|}{I2DFormer+ \textcolor{gray}{[IJCV'24]} \cite{naeem2024i2dformer+}} & ViT
      & 45.9 & 38.3 & 55.2  & 45.3  
      & -- & --  & --  & --  
      & 77.3 & 69.8  & 83.2  & 75.9  \\

      & \multicolumn{2}{r|}{DSECN \textcolor{gray}{[CVPR'24]} \cite{li2024improving}} & ResNet
      & 40.9 & --  & -- & 45.3  
      & 40.0 & --  & --  & 38.5 
      & 49.1 & --  & -- & 53.7  \\

      & \multicolumn{2}{r|}{ZSLViT \textcolor{gray}{[CVPR'24]} \cite{chen2024progressive}} & ViT
      & 78.9 & 69.4  & 78.2  & 73.6  
      & 68.3 & 45.9  & 48.4  & 47.3  
      & 70.7 & 66.1  & 84.6  & 74.2  \\

      & \multicolumn{2}{r|}{PSVMA+ \textcolor{gray}{[TPAMI'24]} \cite{liu2024psvma+}} & ViT
      & 78.8 & 71.8  & 77.8  & 74.6  
      & 74.5 & \underline{61.5} & 49.4  & 54.8  
      & 79.2 & 74.2  & 86.4   & \underline{79.8}  \\

      & \multicolumn{2}{r|}{ZeroMamba \textcolor{gray}{[AAAI'25]} \cite{hou2024zeromamba}} & VMamba 
      & 80.8 & 72.1  & 76.4  & 74.2  
      & 72.4 & 56.5 & 41.4  & 47.7  
      & 71.9 & 67.9  & 87.6   & 76.5  \\

      & \multicolumn{2}{r|}{AENet \textcolor{gray}{[AAAI'25]} \cite{liu2025attend}} & ViT
      & 80.3 & 73.1  & 76.4  & 74.7  
      & 70.0 & 58.6 & 45.2  & 51.0  
      & 75.2 & 70.3  & 80.1   & 74.9  \\

      & \multicolumn{2}{r|}{VSPCN \textcolor{gray}{[CVPR'25]} \cite{jiang2025visual}} & ViT 
      & 80.6 & 72.8 & 78.9 & \underline{75.7}  
      & 75.3 & 59.4 & 49.1 & 53.8  
      & 76.6 & 71.8 & 84.3 & 77.6 \\
    \cdashline{1-16}

    \multirow{15}[2]{*}{\rotatebox{90}{Generative}}
      & \multicolumn{2}{r|}{HSVA \textcolor{gray}{[NeurIPS'21]} \cite{chen2021hsva}} & ResNet
      & -- & 52.7 & 58.3 & 55.3  
      & -- & 48.6 & 39.0  & 43.3  
      & -- & 56.7 & 79.8  & 66.3 \\

      & \multicolumn{2}{r|}{CE-GZSL \textcolor{gray}{[CVPR'21]} \cite{han2021contrastive}} & ResNet
      & 77.5 & 63.9 & 66.8 & 65.3  
      & 63.3 & 48.8 & 38.6  & 43.1  
      & 70.4 & 63.1 & 78.6  & 70.0  \\

      & \multicolumn{2}{r|}{SC-EGG \textcolor{gray}{[IJCAI'22]} \cite{hong2022semantic}} & ResNet 
      & 75.1 & 64.1 & 73.6 & 68.5  
      & 69.2 & 45.1 & 43.6 & 44.3  
      & 78.2 & 60.9 & 89.3 & 72.4  \\

      & \multicolumn{2}{r|}{VGSE-APN \textcolor{gray}{[CVPR'22]} \cite{xu2022vgse}} & ResNet  
      & 28.9 & 21.9 & 45.5 & 29.5  
      & 38.1 & 24.1 & 31.8 & 27.4  
      & 64.0 & 51.2 & 81.8 & 63.0  \\

      & \multicolumn{2}{r|}{ICCE \textcolor{gray}{[CVPR'22]} \cite{kong2022compactness}} & ResNet  
      & 72.7  & 67.3 & 65.5 & 66.4  
      & -- & -- & --  & --  
      & 78.4 & 65.3 & 82.3 & 72.8  \\

      & \multicolumn{2}{r|}{FREE + ESZSL \textcolor{gray}{[ICLR'22]} \cite{cetin2022closed}} & ResNet
      & -- & 51.6 & 60.4 & 55.7  
      & -- & 48.2 & 36.5 & 41.5  
      & -- & 51.3 & 78.0 & 61.8  \\

      & \multicolumn{2}{r|}{TDCSS \textcolor{gray}{[CVPR'22]} \cite{feng2022non}} & ResNet 
      & -- & 44.2 & 62.8&  51.9 
      & -- & -- & -- & -- 
      & --& 59.2 & 74.9& 66.1\\

      & \multicolumn{2}{r|}{DSP \textcolor{gray}{[ICML'23]} \cite{chen2023evolving}} & ResNet 
      & -- & 51.4 & 63.8 & 56.9  
      & -- & 48.3 & 43.0 & 45.5  
      & -- & 60.0 & 86.0 & 70.7  \\

      & \multicolumn{2}{r|}{EGANS \textcolor{gray}{[TEVC'23]} \cite{chen2023egans}} & ResNet 
      & 60.2 & 47.8 &59.2 &52.9  
      & 62.8 & 44.2& 37.4 &40.5
      & 70.6 & 53.9 &81.8 & 65.0\\

      &\multicolumn{2}{r|}{ZeroNAS \textcolor{gray}{[TPAMI'23]} \cite{yan2021zeronas}} & ResNet 
      & 66.4 &56.0 & 63.8 & 59.6  
      & 68.3 & 47.1 & 41.8 & 44.3 
      & 73.2 & 61.4 & 75.3 & 67.6\\

      & \multicolumn{2}{r|}{CDL + OSCO \textcolor{gray}{[TPAMI'23]} \cite{cavazza2023no}} & ResNet
      & -- & 29.0 & 69.0 & 40.6  
      & -- & 32.0  & \textbf{65.0} & 42.9
      & -- & 48.0 & 71.0 & 57.1  \\

      & \multicolumn{2}{r|}{\footnotesize TF-VAEGAN + SHIP$^\star$ \textcolor{gray}{[ICCV'23]}\cite{wang2023improving}} &  ViT
      & -- & 21.1 & \textbf{84.4}  & 34.0  
      & -- & --  & -- & --  
      & -- & 43.7 & \textbf{96.3} & 60.1  \\

      & \multicolumn{2}{r|}{VADS \textcolor{gray}{[CVPR'24]} \cite{Hou_2024_CVPR}} & ViT 
      & \underline{86.8} & \underline{74.1} & 74.6 & 74.3 
      & \underline{76.3} & \textbf{64.6}  & 49.0  & \underline{55.7}  
      & {\underline{82.5}} & \underline{75.4}  & 83.6  & 79.3  \\

      & \multicolumn{2}{r|}{ViFR \textcolor{gray}{[IJCV'25]} \cite{chen2025semantics}} & ResNet 
      & 74.5 & 63.9 & 72.0 & 67.6  
      & 69.2 & 51.3 & 40.0 & 44.7  
      & 77.8 & 68.2 & 78.9 & 73.2 \\

      % & \multicolumn{2}{r|}{GenZSL$^\star$ \textcolor{gray}{[ICML'25]} \cite{chen2025genzsl}} & ViT 
      % & -- & 53.9 & 61.9 & 57.4  
      % & 73.5 & 50.6 & 43.8 & 47.0  
      % & \textbf{92.2} & \textbf{86.1} & 88.7 & \textbf{87.4} \\
    \cmidrule{2-16}

     & \multicolumn{2}{r|}{\cellcolor{tabhighlight}RLVC (ours)} & \multicolumn{1}{c|}{\cellcolor{tabhighlight} ViT} 
     & \cellcolor{tabhighlight}{\textbf{90.1}} & \cellcolor{tabhighlight}\textbf{80.9}  & \cellcolor{tabhighlight}\underline{81.4}  & \cellcolor{tabhighlight}\textbf{81.2}  
     & \cellcolor{tabhighlight}{\textbf{77.7}} & \cellcolor{tabhighlight}59.6 & \cellcolor{tabhighlight}\underline{55.6}  & \cellcolor{tabhighlight}\textbf{57.6}  
     & \cellcolor{tabhighlight}{\textbf{84.0}} & \cellcolor{tabhighlight}\textbf{78.4} & \cellcolor{tabhighlight}{82.4} & \cellcolor{tabhighlight}\textbf{80.4}  \\
    \bottomrule
  \end{tabular}
  }
  \vspace{-3mm}
\end{table*}

\subsection{Experimental Setup}
\label{experiment_setup}
\noindent{\bf Datasets.}
We evaluate on three widely used ZSL datasets:
\begin{itemize}
    \item {\bf CUB} \cite{Welinder2010CaltechUCSDB2}: a fine-grained bird dataset with 11{,}788 images over 150/50 seen/unseen classes and 312 attributes.
    \item {\bf SUN} \cite{patterson2012sun}: a fine-grained scene corpus totals 14{,}340 images from 645/72 seen/unseen classes with 102 attributes.
    \item {\bf AWA2} \cite{xian2019f}: a coarse-grained animal dataset comprising 37{,}322 images across 40/10 seen/unseen classes with 85 attribute annotations.
\end{itemize}

\noindent{\bf Evaluation Metric.}
Following the standard evaluation protocol \cite{chen2024progressive,chen2024causal}, we evaluate RLVC under both CZSL and GZSL with average top-1 accuracy. In the CZSL setting, we report the accuracy on the test set of unseen classes (\textbf{Acc}).
In the GZSL setting, we report accuracy on the seen (\textbf{S}) and unseen (\textbf{U}) test sets, together with their harmonic mean
\textbf{H $=$ (2 $\times$ S $\times$ U) / (S $+$ U)}, which balances performance across the two splits.

\noindent{\bf Implementation Details.}
To ensure a fair comparison and reproducibility, we provide implementation details.
For the visual encoder, we fine-tune a ViT to alleviate cross-dataset bias and use the \texttt{[CLS]} token to extract the visual feature \cite{oquab2023dinov2}.
We adopt the Adam optimizer (betas $=(0.5,\,0.999)$) \cite{kingma2014adam}. The learning rates are $5\times10^{-4}$ for Eq.~\eqref{eq:gen-adv} and $5\times10^{-5}$ for Eq.~\eqref{eq:rl-loss}. 
We activate RL at $E_{\mathrm{RL}}~{=}~30$ for CUB and SUN, and at $E_{\mathrm{RL}}~{=}~7$ for AWA2. 
The prototype-distillation weight $\lambda_{\mathrm{PD}}$, the number of synthesized samples per class, and the total number of training epochs are set to
$\{20,\,1,\,5\}$, $\{400,\,400,\,4000\}$, and $\{500,\,300,\,30\}$ for CUB, SUN, and AWA2, respectively.
All the experiments are run on a single NVIDIA RTX 4090 GPU (24 GB) and implemented using the PyTorch framework.

\begin{table*}[!t]
  \caption{
   Effectiveness validation of RLVC across different semantic prototypes, including word embeddings of class names and expert-annotated attribute vectors.
   We mark the best results in \textbf{bold} and the accuracy gains (\%) in \textcolor{red}{parentheses}.
  }
  \label{tab:semantic_prototype}
  \centering
  \resizebox{1.0\textwidth}{!}{
  \begin{tabular}{l|c|cccc|cccc|cccc}
    \toprule
    \multirow{2}[2]{*}{\textbf{Method}} & 
    \multicolumn{1}{c|}{\multirow{2}[2]{*}{\textbf{Semantic prototype}}} & 
    \multicolumn{4}{c|}{CUB} & 
    \multicolumn{4}{c|}{SUN} & 
    \multicolumn{4}{c}{AWA2} \\
    \cmidrule(rl){3-6} \cmidrule(rl){7-10} \cmidrule(rl){11-14}
    & & 
    \textbf{Acc} & \textbf{U} & \textbf{S} & \textbf{H} & 
    \textbf{Acc} & \textbf{U} & \textbf{S} & \textbf{H} & 
    \textbf{Acc} & \textbf{U} & \textbf{S} & \textbf{H} \\
    \midrule

      Vanilla model & word embedding &61.8&52.4&66.3&58.5&72.0 & 55.4	&46.4&	50.5&73.1 &62.8&82.3&71.2	\\

      \cellcolor{tabhighlight}RLVC &\cellcolor{tabhighlight} word embedding& \cellcolor{tabhighlight}\textbf{62.8} \textcolor{red}{(+1.0)}&\cellcolor{tabhighlight}51.5&\cellcolor{tabhighlight}70.9&\cellcolor{tabhighlight}\textbf{59.7} \textcolor{red}{(+1.2)}&\cellcolor{tabhighlight}\textbf{72.1} \textcolor{red}{(+0.1)}&\cellcolor{tabhighlight}55.0&\cellcolor{tabhighlight}50.3	&\cellcolor{tabhighlight}\textbf{52.5} \textcolor{red}{(+2.0)}&\cellcolor{tabhighlight}\textbf{74.5} \textcolor{red}{(+1.4)}&\cellcolor{tabhighlight}61.3	&\cellcolor{tabhighlight}91.7&	\cellcolor{tabhighlight}\textbf{73.5} \textcolor{red}{(+2.3)}	\\
      
    \cdashline{1-14}
      Vanilla model  & attribute vector
     & 88.6  & 71.0	&79.8	& 75.1 &75.8 & 58.3	&52.3 & 55.1 & 75.7 &70.1	& 76.0 & 72.8 \\
      \cellcolor{tabhighlight}RLVC & \cellcolor{tabhighlight}attribute vector      & \cellcolor{tabhighlight}{\textbf{90.1}} \textcolor{red}{(+1.5)} & \cellcolor{tabhighlight}80.9  & \cellcolor{tabhighlight}81.4  & \cellcolor{tabhighlight}\textbf{81.2} \textcolor{red}{(+6.1)}
     & \cellcolor{tabhighlight}{\textbf{77.7}} \textcolor{red}{(+1.9)} & \cellcolor{tabhighlight}59.6 & \cellcolor{tabhighlight}55.6  & \cellcolor{tabhighlight}\textbf{57.6} \textcolor{red}{(+2.6)}
     & \cellcolor{tabhighlight}{\textbf{84.0}} \textcolor{red}{(+8.3)}& \cellcolor{tabhighlight}78.4 & \cellcolor{tabhighlight}{82.4}  & \cellcolor{tabhighlight}\textbf{80.4} \textcolor{red}{(+7.6)}\\
    \bottomrule
  \end{tabular}
  }
\end{table*}

\subsection{Core Results}
Table \ref{tab:gzsl} compares our RLVC with recent SOTA embedding and generative ZSL methods reported in top-tier venues. 
We highlight our findings below: 
\textbf{(1) RLVC exhibits the best CZSL accuracy on all three benchmarks}, \ie, 90.1\%, 77.7\%, and 84.0\% on CUB, SUN, and AWA2. These accuracies surpass the previous best method, VADS \cite{Hou_2024_CVPR}, by 3.1\%, 1.4\%, and 1.5\%, respectively.
\textbf{(2) In the GZSL setting, RLVC achieves the top harmonic mean H on all datasets and strikes a more effective trade-off between seen and unseen accuracy.}
Compared with the second-best methods, like VSPCN \cite{jiang2025visual} on CUB, VADS \cite{Hou_2024_CVPR} on SUN, and PSVMA+ \cite{liu2024psvma+} on AWA2, RLVC notably increases \textbf{H} by 5.5\%, 1.5\%, and 0.6\%, respectively. 
While RLVC does not achieve the best values of \textbf{S} and \textbf{U} on every dataset, it balances them more effectively, resulting in the highest \textbf{H} across all datasets. 
Both \textbf{U} and \textbf{S} are competitive against SOTA methods using  ViT, VMamba, and ResNet backbones.
Moreover, the VLM-based method SHIP \cite{wang2023improving} demonstrates outstanding performance on seen classes, yet it fails to generalize well to unseen classes.

These results consistently demonstrate the superiority and strong generalization of our RLVC. Furthermore, they indicate that the proposed outcome-reward RL with visual cues effectively guides generation toward task-relevant features rather than merely distributionally plausible ones.

\subsection{RLVC with Different Semantic Prototypes}
To verify the reliability and show the robustness of our proposed RLVC, particularly for semantically similar classes, we conduct a comparative experiment and analysis across different semantic prototypes.
We consider two types of prototypes: word embeddings of class names from the CLIP text encoder \cite{radford2021learning} and expert-annotated attribute vectors. 
Maintaining the same hyperparameters, we compare the performance of the full RLVC to that of a vanilla generative model (\ie, without the proposed RL and visual cues).

As shown in Table \ref{tab:semantic_prototype}, empirical results indicate that RLVC improves both \textbf{Acc} and \textbf{H} across all class prototype types and all datasets, with gains ranging from 0.1\% to 8.3\%.
These results confirm that RLVC provides a faithful and task-relevant feature synthesis. Consequently, it enhances the discriminative power of visual features, even when using challenging word embeddings of class names that encode strong semantic similarity.
To some extent, this accuracy increase also highlights the importance of aligning synthesized features with visual prototypes.
It is worth noting that our RLVC also surpasses CLIP on CUB.

\subsection{Ablation Study and Analysis}
	\begin{table}[t]
		\centering
		\caption{Results of RLVC variants on CUB, SUN and AWA2 datasets. We ablate specific components to assess their effectiveness.
        The \textbf{bold} marking indicates the best results.} \label{table:ablation}
		\resizebox{0.49\textwidth}{!}
		{
			\begin{tabular}{l|cc|cc|cc}
            \toprule
				\multirow{2}*{\textbf{Configuration}} &\multicolumn{2}{c|}{CUB} &\multicolumn{2}{c|}{SUN}&\multicolumn{2}{c}{AWA2} \\
				\cmidrule(rl){2-3}\cmidrule(rl){4-5}\cmidrule(rl){6-7}
                %\cline{2-3}\cline{4-5}\cline{6-7}
				&\textbf{Acc}& \textbf{H} &\textbf{Acc}& \textbf{H} &\textbf{Acc}& \textbf{H}\\
                \midrule
                RLVC w/o RL \& visual cues 
                & 88.6 & 75.1
                & 75.8 & 55.1
                & 75.7 & 72.8\\
                RLVC w/o RL (\ie, Eq. \eqref{eq:rl-loss})
                & 89.2 & 80.1
                & 76.1 & 55.6
                & 79.4 & 73.9\\
                RLVC w/o visual cues (\ie, Eq. \eqref{eq:pd})
                & 88.9 & 79.2
                & 77.0 & 56.9
                & 74.9 & 76.6\\
                RLVC w/o fine-tuning visual encoder
                & 89.2 & 77.5
                & 76.0 & 56.4
                & 81.1 & 76.3\\
                RLVC w/o advantage (\ie, Eq. \eqref{eq:adv})
                & 89.6 & 79.7
                & 76.0 & 56.0
                & 82.4 & 78.2\\
				\cdashline{1-7}
                \cellcolor{tabhighlight}RLVC  & \cellcolor{tabhighlight}\textbf{90.1} & \cellcolor{tabhighlight}\textbf{81.2} & \cellcolor{tabhighlight}\textbf{77.7} &  \cellcolor{tabhighlight}\textbf{57.6} & \cellcolor{tabhighlight}\textbf{84.0} & \cellcolor{tabhighlight}\textbf{80.4}\\
				\bottomrule
			\end{tabular}
		}
	\end{table}
\noindent {\bf Ablation on Components.}
In Table \ref{table:ablation}, we conduct comprehensive ablation studies to evaluate the effectiveness of our proposed designs in RLVC.
Applying our proposed operations results in a significant improvement in accuracy over the vanilla model.
In CZSL, the average increase in \textbf{Acc} is 3.9\%, and in GZSL, the average increase in \textbf{H} is 5.3\%.
{\bf RL is crucial:} Without RL, the performance drops significantly, especially on the AWA2 dataset (\eg, \textbf{Acc}/\textbf{H} drops from 84.0\%/80.4\% to 79.4\%/73.9\%). This indicates that RL significantly enhances the model's learning capabilities in both data distribution and task relevance.
{\bf Visual cues are beneficial:} Mining visual cues as visual prototypes and applying a prototype-distillation loss centralizes the synthesized features while also stabilizing training. It can be observed that omitting this component substantially degrades performance on all benchmarks.
{\bf Visual encoder fine-tuning is essential:} We do not optimize the visual encoder separately. Instead, we fine-tune it with the reward model. This reduces training overhead and simultaneously alleviates domain bias. The fine-tuned visual features inject dataset-specific priors, which is beneficial for GZSL (\textbf{H} increases by 3.7\%, 1.2\%, and 4.1\% for CUB, SUN, and AWA2, respectively).
{\bf EMA reward smoothing outperforms raw reward:} Our investigation demonstrates that smoothing the reward using Eq. \eqref{eq:adv} offers substantial performance advantages over using the raw reward directly. This also suggests that algorithms specifically tailored for RL optimization are beneficial.

\noindent {\bf Ablation on Prototype-Distillation Losses.}
	\begin{table}[t]
		\centering
		\caption{ Comparison results for different prototype-distillation losses combined with RLVC on CUB, SUN and AWA2 datasets. The \textbf{bold} marking indicates the best results.} \label{table:ablation_loss}
		\resizebox{0.45\textwidth}{!}
		{
			\begin{tabular}{l|cc|cc|cc}
            \toprule
				\multirow{2}*{\textbf{Loss}} &\multicolumn{2}{c|}{CUB} &\multicolumn{2}{c|}{SUN}&\multicolumn{2}{c}{AWA2} \\
				\cmidrule(rl){2-3}\cmidrule(rl){4-5}\cmidrule(rl){6-7}
                %\cline{2-3}\cline{4-5}\cline{6-7}
				&\textbf{Acc}& \textbf{H} &\textbf{Acc}& \textbf{H} &\textbf{Acc}& \textbf{H}\\
				\midrule
				$\rm KL$ & 88.9 & 80.1& 77.2 & \textbf{58.2} & 76.4 & 76.4 \\
                $\ell_{\rm{1}}$ & 89.2 & 80.9 & 77.3 & 57.8 & 75.9 & 77.6 \\
                \cellcolor{tabhighlight}$\mathcal{L}_{\rm{PD}}$ & \cellcolor{tabhighlight}\textbf{90.1} & \cellcolor{tabhighlight}\textbf{81.2} & \cellcolor{tabhighlight}\textbf{77.7} &  \cellcolor{tabhighlight}57.6 & \cellcolor{tabhighlight}\textbf{84.0} & \cellcolor{tabhighlight}\textbf{80.4}\\
				\bottomrule
			\end{tabular}
		}
	\end{table}
Compared to the standard KL loss, we introduce a novel prototype-distillation loss (Eq. \eqref{eq:pd}).
As shown in Table~\ref{table:ablation_loss}, our loss outperforms KL and $\ell_1$ in most cases, with the exception of the GZSL setting on SUN.
We attribute this to the proposed loss being better suited for clustering and for distilling prototype information into the model.

\begin{figure*}[!t]
    \centering
    \includegraphics[width=0.95\linewidth]{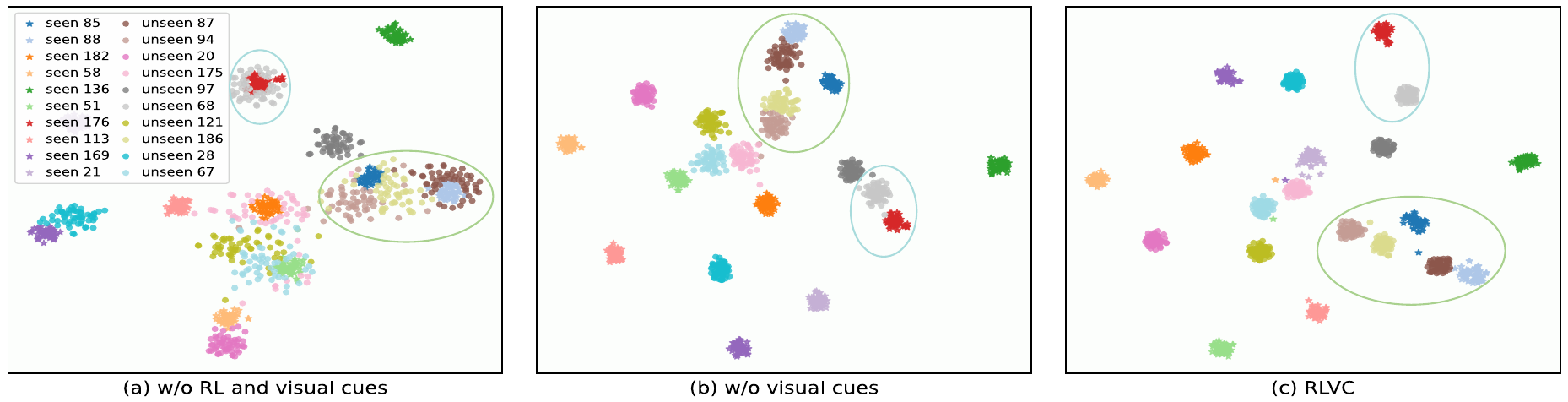}
    \vspace{-3mm}
    \caption{Qualitative t-SNE visualization of RLVC on CUB: (a) without RL and visual cues, (b) without visual cues, and (c) full RLVC. We use real features of seen classes and synthetic features of unseen classes. Zoom in for details.}%More visualization on SUN and AWA2 in the Appendix. Zoom in for details.}
    \label{fig:tsne_cub}
    \vspace{-2mm}
\end{figure*}

\begin{figure*}[!t]
  \centering
  \begin{subfigure}[t]{0.33\linewidth}
    \centering
    \includegraphics[width=\linewidth]{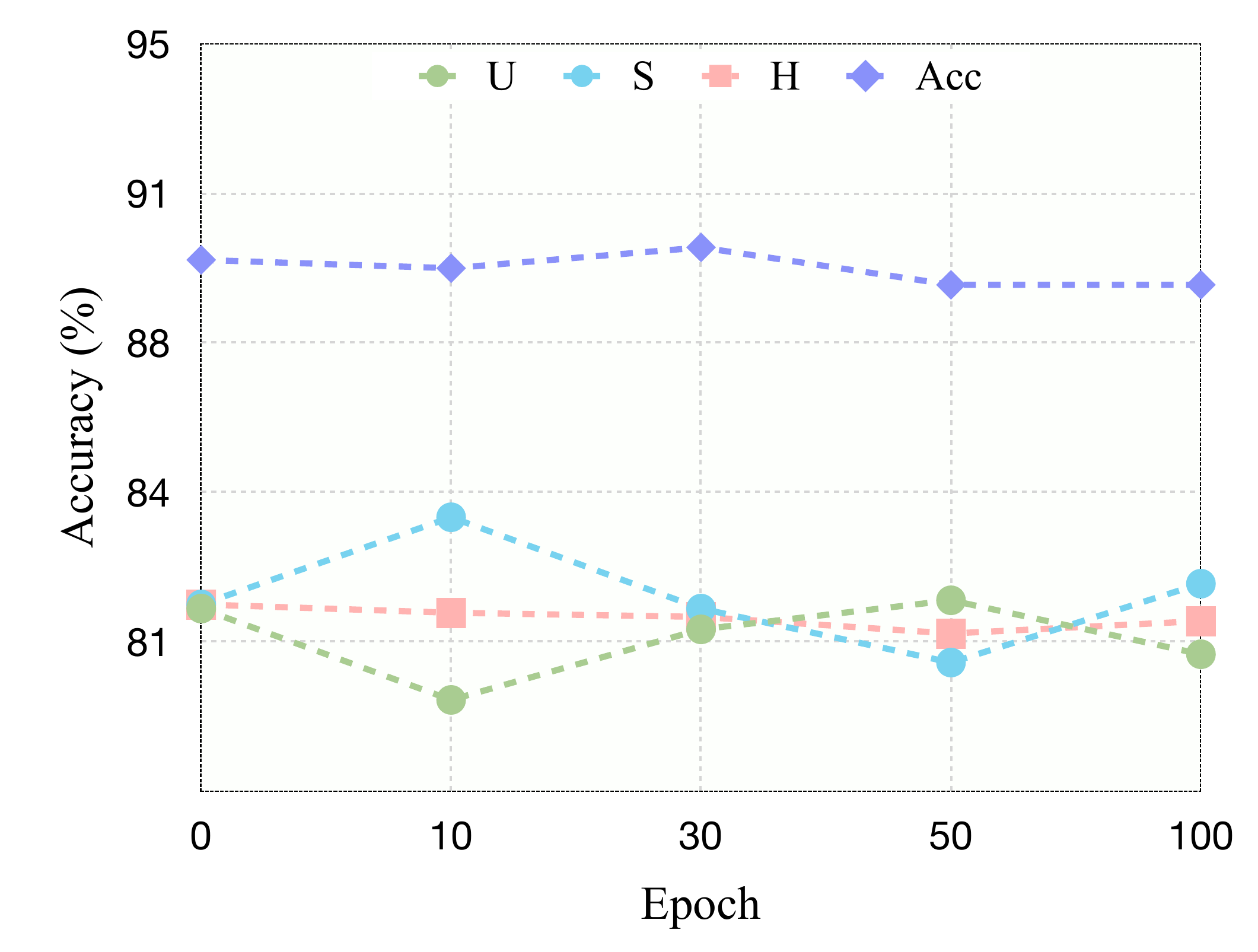} % fig1.pdf/png/eps
    %\caption{a}
    \label{fig:sub1}
  \end{subfigure} \hspace{-3mm}
  \begin{subfigure}[t]{0.33\linewidth}
    \centering
    \includegraphics[width=\linewidth]{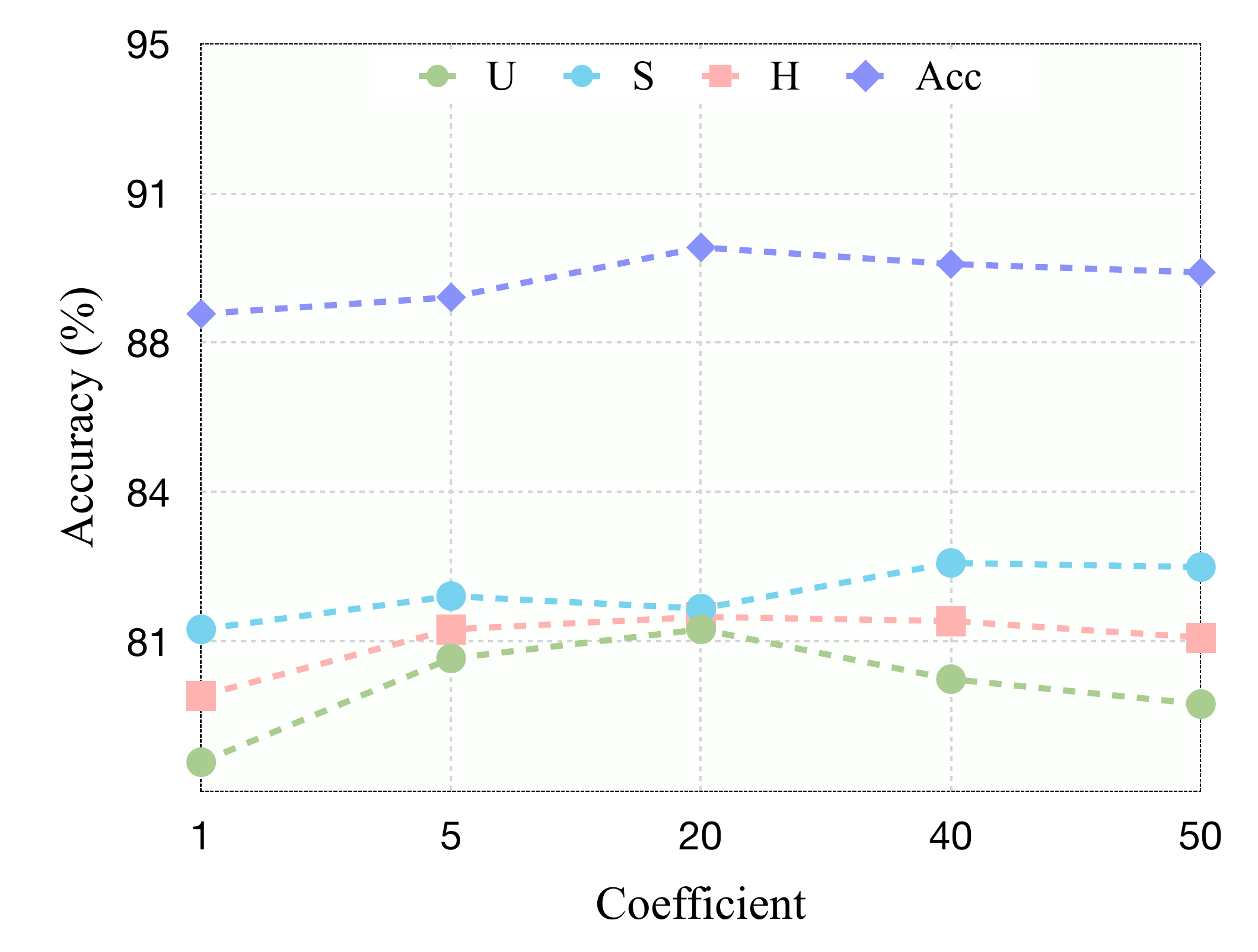}
    %\caption{b}
    \label{fig:sub2}
  \end{subfigure} \hspace{-3mm}
  \begin{subfigure}[t]{0.33\linewidth}
    \centering
    \includegraphics[width=\linewidth]{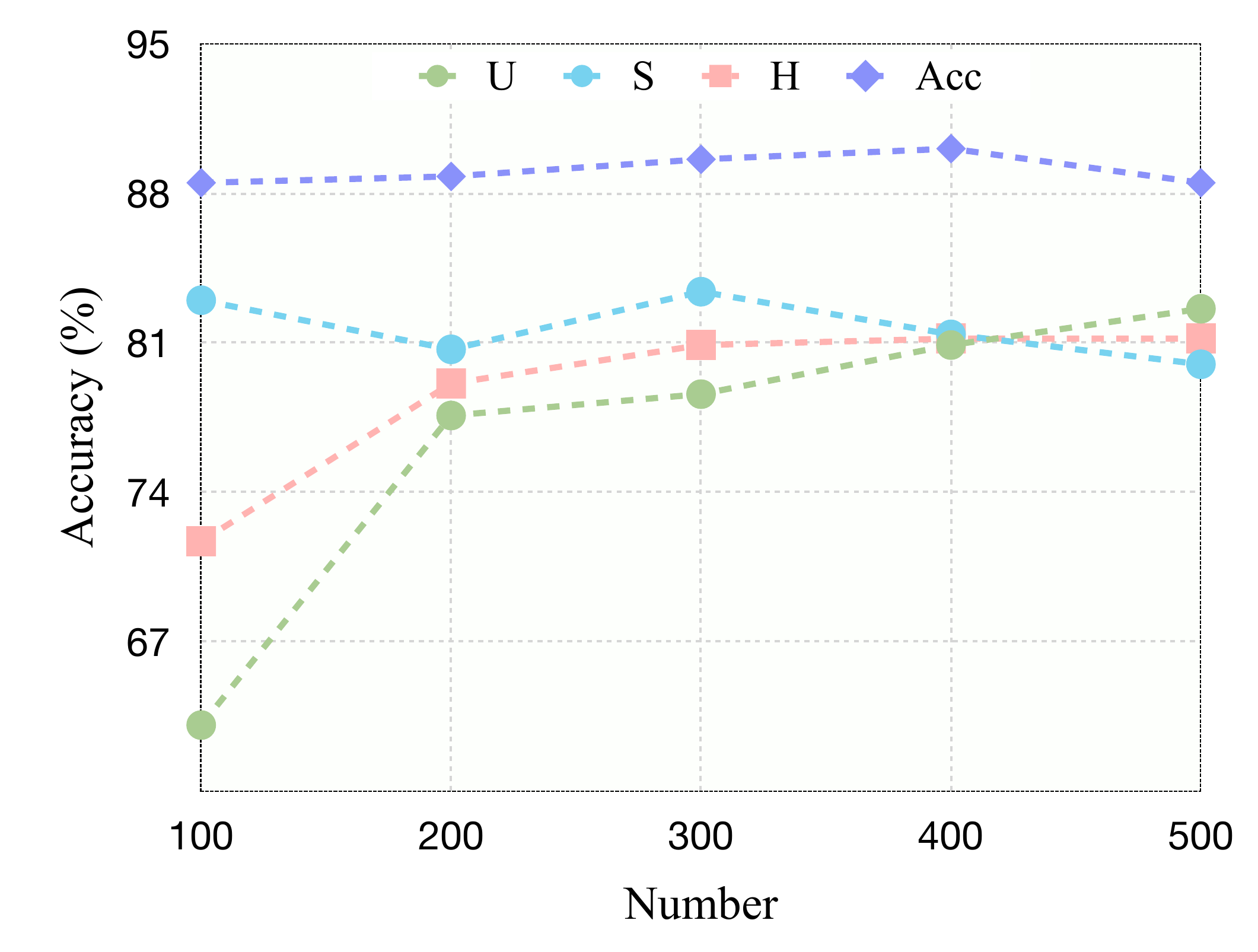}
    %\caption{c}
    \label{fig:sub3}
  \end{subfigure}
\vspace{-8mm}
% \caption{Impact of RL cold-start, visual loss coefficient, and number of synthetic unseen samples on CUB.}
\caption{Effect of hyperparameters on CUB, including the epoch of RL cold-start, the coefficient of visual loss, and the number of synthetic unseen samples.}%Results on SUN are provided in the Appendix \textcolor{cvprblue}{C}.} %\ref{app:hyper}.}

  \label{fig:hyper_cub}
\vspace{-2mm}
\end{figure*}

\subsection{Qualitative Evaluation}
To intuitively illustrate the distribution of the synthesized data, we use t-SNE visualization \cite{van2008visualizing} to display the results from (a) the vanilla model (\ie, without RL and visual cues), (b) the model without visual cues, and (c) our full RLVC on CUB in Fig. \ref{fig:tsne_cub}. 
The visualization includes 10 real seen classes and 10 synthesized unseen classes, denoted by $\star$ and $\circ$, respectively.
The \textcolor{tsne_blue}{blue circle} denotes ``Balitimore Oriole" and ``Orchard Oriole", and the \textcolor{tsne_green}{green circle} denotes ``Harris Sparrow", ``Lincoln Sparrow", ``Le Conte Sparrow", ``White crowned Sparrow" and ``Tree Sparrow". Within each circle, they are semantically similar. 

Visually, in (a) the vanilla model (\ie, without RL and visual cues), seen and unseen classes are significantly overlapped, particularly for semantically similar categories. 
In contrast,  (b) without visual cues and (c) our RLVC exhibit clear boundaries, contributing to classification. 
Comparatively, RLVC imposes a visual prototype constraint, leading to each class demonstrating more compact clustering.
Overall, our RLVC learns a task-relevant, more accurate data distribution, which aligns with our motivation. %See Appendix \textcolor{cvprblue}{B} for more results.
%\ref{app:tsne}
% As seen from the results, without our operations (), the synthesized features suffer from: (1) confusion between seen and unseen classes, and (2) a scattered feature manifold of the synthesized unseen classes.
% Compared to the model without visual cues, our RLVC achieves: (1) more compact clusters, and (2) more distinct boundaries.
% These results indicate that RLVC promotes the generation of task-relevant class features, rather than merely optimizing the data distribution. Meanwhile, accurate learning data distribution of seen classes preserves more boundary space for the unseen classes, which is consistent with our motivation. More results in the Appendix \ref{app:tsne}.

\subsection{Hyperparameter Analysis}
We conduct a sensitivity analysis for several key hyperparameters on the CUB to validate our default configuration, with results shown in Fig.~\ref{fig:hyper_cub}. 
This analysis includes: the epoch of RL cold-start, the coefficient of visual loss, and the number of synthetic unseen samples. 
The cold-start mechanism stabilizes initial RL training. Our experiments show that a value of 30 ensures stable training while achieving optimal performance.
The visual loss coefficient balances the generative objective against prototype-based feature clustering.
As its value increases, performance first rises, then falls, peaking at 20.
The number of synthetic samples for unseen classes impacts the accuracy balance between seen and unseen classes. The best \textbf{H} is achieved when it is set to 400.
These results confirm the rationality and robustness of our hyperparameter choices. We provide the detailed settings for each dataset in \S\ref{experiment_setup}.

\section{Conclusion}
In this work, we present RLVC, an outcome-reward reinforcement learning framework with visual cues for generative ZSL. 
This framework aims to learn features well-suited for classification by strengthening the model’s generative capability. 
Equipped with a carefully designed cold-start training strategy, RLVC reliably synthesizes features. 
Through qualitative and quantitative experiments on three ZSL benchmarks, we show that RLVC consistently outperforms existing methods. 
As an initial exploration of reinforcement learning for generative ZSL, we highlight RLVC's potential. This approach may also benefit broader tasks and robust training recipes.

\section*{Acknowledgments}
This work was supported by the National Key R\&D Program of China (2023YFC3305600) and the National Natural Science Foundation of China (92570101).

{
    \small
    \bibliographystyle{ieeenat_fullname}
    \bibliography{main}
}


\end{document}